\documentclass[11pt]{article}

\usepackage[final]{acl}

\usepackage{times}
\usepackage{latexsym}

\usepackage[T1]{fontenc}

\usepackage[utf8]{inputenc}

\usepackage{microtype}

\usepackage{inconsolata}

\usepackage{graphicx}

\usepackage{algorithm}
\usepackage{bm}
\usepackage{booktabs}
\usepackage{amsmath,amsfonts}
\usepackage{array}
\usepackage{multirow}
\usepackage{verbatim}
\usepackage{colortbl}
\usepackage{xcolor}
\usepackage{makecell}
\usepackage{algpseudocode}
\usepackage{tabularx}

\usepackage{makecell}

\title{Breaking the Generator Barrier: Disentangled Representation for \\ Generalizable AI-Text Detection}

\author{
Xiao Pu,
Zepeng Cheng,
Lin Yuan,
Yu Wu,
Xiuli Bi\thanks{Corresponding author} 
\\
 \text{Chongqing University of Posts and Telecommunications, China}
\\
\texttt{puxiao@cqupt.edu.cn,chengzepenv@gmail.com}
\\
\texttt{yuanlin@cqupt.edu.cn,wuyu@cqupt.edu.cn,bixl@cqupt.edu.cn}
}

\begin{document}
\maketitle
\begin{abstract}
As large language models (LLMs) generate text that increasingly resembles human writing, the subtle cues that distinguish AI-generated content from human-written content become increasingly challenging to capture. 
Reliance on generator-specific artifacts is inherently unstable, since new models emerge rapidly and reduce the robustness of such shortcuts. 
This generalizes unseen generators as a central and challenging problem for AI-text detection. 
To tackle this challenge, we propose a progressively structured framework that disentangles AI-detection semantics from generator-aware artifacts. 
This is achieved through a compact latent encoding that encourages semantic minimality, followed by perturbation-based regularization to reduce residual entanglement, and finally a discriminative adaptation stage that aligns representations with task objectives. 
Experiments on MAGE benchmark, covering 20 representative LLMs across 7 categories, demonstrate consistent improvements over state-of-the-art methods, achieving up to 24.2\% accuracy gain and 26.2\% $F_1$ improvement. 
Notably, performance continues to improve as the diversity of training generators increases, confirming strong scalability and generalization in open-set scenarios. Our source code will be publicly available at \texttt{\href{https://github.com/PuXiao06/DRGD}{https://github.com/PuXiao06/DRGD.}}
\end{abstract}

\section{Introduction}
The rapid proliferation of large language models (LLMs) has resulted in a surge of AI-generated text (AIGT) across news media, social platforms, and academic domains.
While this development offers new opportunities, it has also raised increasing concerns regarding misinformation~\cite{pu2025dear,wei2025cross,hu2024bad}, manipulation~\cite{guan2024adversarial,shao2023detecting}, and authorship integrity~\cite{silva2024forged,vasilatos2023howkgpt}. 
These concerns have made detecting AI-generated text a pressing task in trustworthy AI and content regulation, especially amid rapidly emerging unseen LLMs.

\begin{figure}
\begin{center}
\includegraphics[scale=.22]{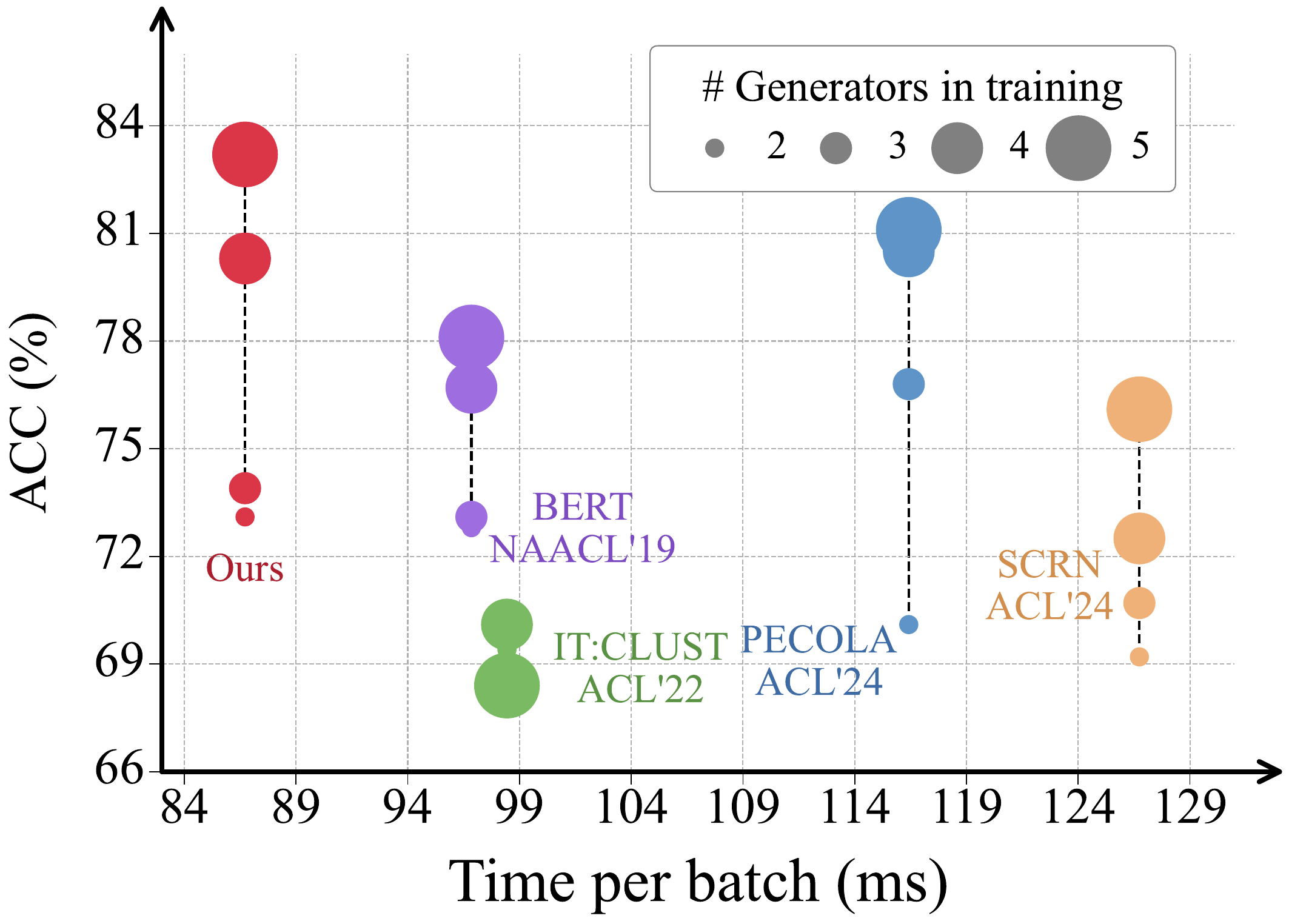}
\end{center}
\caption{Comparison with competitive approaches. 
On the unseen OPT generators, our method achieves the best balance between accuracy and efficiency, and its advantage grows with training generator diversity.}
\label{fig:intro}
\end{figure}

Although recent advances in AIGT detection have shown promising results under closed-world settings where training and test samples share the same generator~\cite{wang2025novel,liu2024does}, these methods fail to generalize to previously unseen generators, resulting in substantial performance degradation.
This issue is especially pronounced in real-world settings, where novel LLMs are continuously emerging.
While prior efforts have explored cross-domain generalization through adversarial~\cite{huang2024ai} or contrastive learning~\cite{liu2024does}, they tend to treat generator shift as a subset of domain shift and often overlook its distinct challenges. 
Recent studies~\cite{wu2025survey,guo2024biscope,li2024mage,wang2024m4} have shown that shifts across generators introduce unique semantic, syntactic, and stylistic variations, resulting in generalization gaps beyond the scope of domain-centric solutions.

To address the core challenge of generalization to unseen LLM-based generators, we propose a progressively structured framework that disentangles AI-detection semantics from generator-aware artifacts.
The goal is to isolate task-relevant signals by suppressing entangled generator bias.
The process begins with a dual-bottleneck design that enforces compactness and minimality in the latent space, encouraging essential task-focused representations.
Building on this, cross-view regularization disrupts residual correlations and promotes independence among factors.
Finally, a discriminator-guided adaptation stage further consolidates the separation by refining each representation stream against task-specific objectives and suppressing residual leakage across branches.
This progressive pipeline incrementally purifies AI-detection semantics while filtering out generator-aware noise, yielding representations that generalize more effectively across diverse LLMs.

In summary, our contributions are as follows:
\begin{itemize}
    \item We address the challenge of AIGT detection on unseen LLMs by learning task-focused representations that isolate AI-detection cues from generator-aware artifacts, enabling stronger generalization in open-set scenarios.
    \item  
    Our solution follows a progressive design: it begins with dual-bottleneck encoding to encourage semantic compactness, applies cross-view regularization to disrupt entangled factors, and finalizes with discriminative-guided adaptation that sharpens task alignment and reinforces representational separation.
    \item Experiments on 20 representative LLMs demonstrate consistent improvements over state-of-the-art methods, with increasingly larger gains under diverse generator settings, validating the effectiveness and generalizability of our approach (see Figure~\ref{fig:intro}).
\end{itemize}

\begin{figure*}[t]
    \centering
    \includegraphics[scale=0.6]{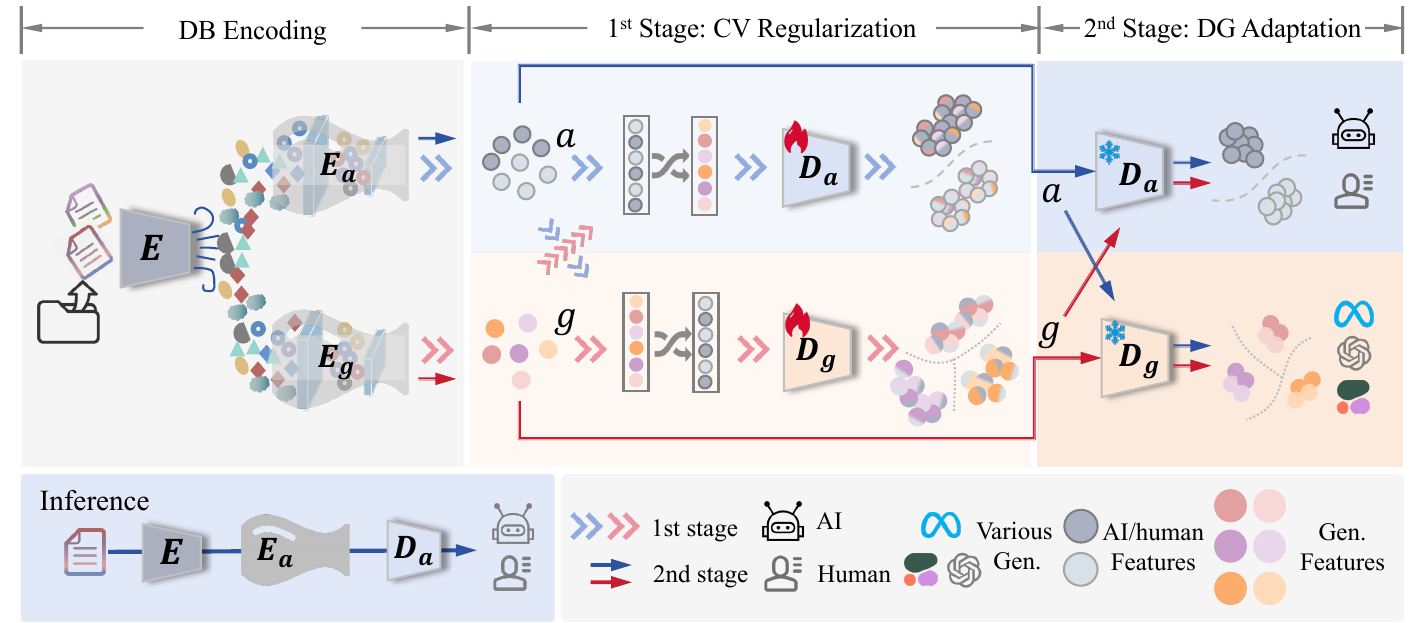} 
     \caption{Overview of the proposed framework. The model enhances generalization to unseen generators by enforcing semantic disentanglement between AI-detection and generator-aware features. It integrates compact dual-bottleneck (DB) encoding, cross-view (CV) regularization, and discriminator-guided (DG) adaptation into a cohesive disentanglement-enhancing pipeline.}
     \label{fig:main}
\end{figure*}
\section{Related Work}
\subsection{AI-generated Text Detection}
With the widespread adoption of large language models (LLMs) such as GPT-3~\cite{brown2020language}, GPT-4~\cite{achiam2023gpt}, and Claude~\cite{anthropic2026claude46}, detecting AIGT has become a central task in trustworthy NLP. 
Existing detection methods can be categorized into four main groups.

Statistic-based approaches~\cite{zhou2025adadetectgpt,shi2024ten,su2023detectllm,gehrmann2019gltr} rely on entropy, perplexity, or curvature-based signals to distinguish human-written text from machine-generated outputs. 
Watermark-based methods~\cite{liu2024an,kirchenbauer2023watermark} inject identifiable patterns into the generation process to facilitate detection. 
Classifier-based techniques~\cite{liu2024does,hu2023radar, shnarch2022cluster,uchendu2020authorship} typically fine-tune pretrained models such as RoBERTa or BERT, often incorporating adversarial or contrastive objectives. 
Retrieval-based approaches ~\cite{krishna2023paraphrasing} match text against known generator outputs via semantic similarity.

Although many existing approaches achieve strong performance in closed-world settings, where training and test data originate from the same generator, they often fail to generalize to previously unseen generators. 
Recent benchmarks such as MAGE~\cite{li2024mage} and M4~\cite{wang2024m4} have systematically exposed such weaknesses, revealing substantial performance drops under generator shifts. 
These findings underscore the need for open-set detectors that explicitly model generator-aware variations.

\subsection{Beyond Domain Generalization: Generator-Aware Challenges}
To improve robustness under distribution shifts, several studies have explored domain generalization (DG) techniques, which were initially developed for cross-topic or cross-genre transfer in NLP and vision.  
These methods aim to learn domain-invariant features through adversarial training~\cite{tuck2026guided,li2025iron,ganin2016domain}, feature alignment~\cite{li2018domain}, or data augmentation~\cite{ li2022uncertainty,peng2018sim}.
Motivated by their effectiveness, DG strategies have been adapted for AI text detection~\cite{liu2024does, huang2024ai}, showing modest gains on certain unseen generators.

However, most DG-based approaches implicitly assume that generator shift resembles domain shift. 
Studies such as MAGE~\cite{li2024mage} reveal that generator-aware variation introduces unique stylistic and semantic artifacts that differ qualitatively from conventional domain differences. 
Additional findings~\cite{chen2025imitate, wang2024m4, jiang2023improving} show that generators vary not only in syntax or topical preference, but also in prompt interpretation and semantics, factors that standard domain-invariant methods fail to disentangle.

These limitations highlight the need for approaches that transcend conventional domain generalization by explicitly modeling generator-aware semantics. 
Unlike prior one-shot learning methods that implicitly subsume generator variation under general domain shifts, our approach progressively disentangles detection-relevant semantics from generator-aware artifacts, enabling robust and interpretable generalization to unseen LLMs.

\section{Methodology}\label{sec:method}
We propose a progressively structured framework for detecting AIGT from unseen generators by learning task-specific representations that disentangle AI-detection semantics from generator-aware patterns.
The framework begins with a dual-bottleneck encoding that encourages compact and factorized representations by minimizing shared redundancy. A cross-view regularization then disrupts residual entanglement across branches, enhancing representational independence.
Finally, discriminative adaptation refines each branch using dedicated pre-trained discriminators, aligning each with its target objective while suppressing interference from the opposite one.
Together, these components form a cohesive pipeline that enhances semantic purity and significantly improves generalization across diverse generative sources.
The overall architecture is shown in Figure~\ref{fig:main}.

\subsection{Problem Setup}
We address the task of detecting whether a given text sample is AI-generated, even when the generator that produced it has not been observed during training.
Each training instance is denoted as $\mathcal{X} = \{(x_{i}, y_{i}, s_{i})\}_{i=1}^{N}$, where $x_{i}$ is a text sample, $y_{i} \in \{0, 1\}$ is the source label (human or AI) and $s_{i}$ denotes the generator identity (e.g., GPT, LLaMA).
This formulation enables the model to learn both AI-detection cues and generator-aware variations that are useful for open-set generalization.

\subsection{Dual-Bottleneck Encoding}
To improve generalization under distribution shifts from unseen LLM-based generators, we propose a dual-bottleneck encoding module that disentangles AI-detection semantics from generator-aware artifacts. 
This is achieved by enforcing compact, task-aligned latent representations guided by the information bottleneck (IB) principle.

The IB objective encourages latent features $z$ to retain information about the target label $y$ while discarding irrelevant input variations: $\min \big[-I(z; y) + \beta I(x; z)\big]$, where $\beta$ balances prediction fidelity and compression.
In practice, this translates into a prediction loss (approximating $-I(z;y)$) combined with a KL regularization term (approximating $I(x;z)$).

Concretely, we implement two parallel encoder branches: $E_{a}$ for AI-detection semantics and $E_{g}$ for generator-aware cues.
Given an input $x_{i}$, we first obtain a contextual embedding $\mathbf{h}_{i}$ from a pre-trained BERT [CLS] token.
This is projected into intermediate vectors $\mathbf{e}_{i}^{(a)}$ and $\mathbf{e}_{i}^{(g)}$ via MLPs, which parameterize diagonal-Gaussian posteriors:
\begin{equation}
q(\mathbf{a}_{i} \mid \mathbf{e}_{i}^{(a)}) = \mathcal{N}\bigg(\boldsymbol{\mu}_{i}^{(a)}, \operatorname{diag}\big((\boldsymbol{\sigma}_{i}^{(a)})^2\big)\bigg),
\end{equation}
where \( \boldsymbol{\mu}_i^{(a)} \) and \( \boldsymbol{\sigma}_i^{(a)} \) are computed via separate linear layers, with softplus activation applied to the variance to ensure positivity.
An analogous operation is applied to obtain the generator-aware posterior $q(\mathbf{g}_{i} \mid \mathbf{e}_{i}^{(g)})$.

To enable differentiable sampling during training, we draw $K$ samples using the reparameterization trick~\cite{kingma2013auto}:
\begin{equation}
\mathbf{a}_{i}^{(K)} = \boldsymbol{\mu}_{i}^{(a)} + \boldsymbol{\sigma}_{i}^{(a)} \odot \boldsymbol{\epsilon}^{(K)}, \quad \boldsymbol{\epsilon}^{(K)} \sim \mathcal{N}(0, \mathbf{I}),
\end{equation}
where \( \odot \) denotes element-wise multiplication.
Then we average the prediction losses across $K$ Monte Carlo samples for stability.
At inference time, the latent feature is taken as the posterior mean: $\mathbf{a}_{i} = \boldsymbol{\mu}_{i}^{(a)}$.
The same sampling and mean-inference strategy is applied to derive the latent representation $\mathbf{g}_{i}$.

To enforce compactness and disentanglement, we introduce learnable priors \( p(\mathbf{a}) \) and \( p(\mathbf{g}) \) for each branch, which is initialized as a standard Gaussian $\mathcal{N}(0, \mathbf{I})$ and optionally updated during training. 
We then apply KL regularization as follows:
\begin{align}
\mathcal{L}_{\text{DB}} ={} & D_{\text{KL}} \left( q(\mathbf{a}_{i} \mid \mathbf{e}_{i}^{(a)}) \,\|\, p(\mathbf{a}) \right) \notag \\
& \hspace{2em} + D_{\text{KL}} \left( q(\mathbf{g}_{i} \mid \mathbf{e}_{i}^{(g)}) \,\|\, p(\mathbf{g}) \right).
\label{eq:ib}
\end{align}
This regularization encourages latent representations $\mathbf{a}_{i}$ and $\mathbf{g}_{i}$ to align with task-agnostic priors, minimizing redundancy while promoting task-specific compression.

This dual-bottleneck design enforces semantic compression and separation at the feature level, providing a solid foundation for subsequent regularization and adaptation modules to operate on disentangled, robust representations.

\subsection{Cross-View Regularization}
While the dual-bottleneck encoding encourages semantic decoupling, residual entanglement may still persist due to implicit correlations and shared linguistic patterns across generators.
To further separate AI-detection and generator-aware semantics, we introduce a cross-view regularization mechanism that explicitly perturbs each representation with signals from its complementary branch, thereby promoting representational independence.

For each input $x_{i}$, we sample another instance $x_j\ne x_i$ ($x_j$ is a human or AI sample) within the batch and perturb the AI-detection feature $\mathbf{a}_{i}$ using the generator-aware representation $\mathbf{g}_{j}$.
The perturbed output $\tilde{\mathbf{a}}_i$ is computed as:
\begin{align}
\tilde{\mathbf{a}}_{i} = \gamma \cdot \mathbf{a}_i + (1 - \gamma) \cdot \phi(\mathbf{g}_{j}, \mathbf{a}_{i}),
\end{align}
where $\gamma \sim \mathcal{U}(0.5, 1)$ controls the interpolation, and $\phi(\cdot, \cdot)$ transfers the statistics style of $\mathbf{g}_i$ to $\mathbf{a}_j$:
\begin{equation}
\phi(\mathbf{g}_{i}, \mathbf{a}_{j}) = \frac{\mathbf{a}_{j} - \mu(\mathbf{a}_{j})}{\sigma(\mathbf{a}_{j})} \cdot \sigma(\mathbf{g}_{i}) + \mu(\mathbf{g}_{i}),
\end{equation}
with $\mu(\cdot)$ and $\sigma(\cdot)$ denoting mean and standard deviation over features.
A symmetric operation is applied to perturb $\tilde{\mathbf{g}}_{i}$ using $\mathbf{a}_{j}$

To further mitigate asymmetry, where AI-generated samples are more susceptible to misclassifications, we apply additional cross-branch perturbations exclusively to AI-generated samples, producing augmented variant $\tilde{\mathbf{a}}_{i}^{(\text{aug})}$.
We then apply a dual-branch prediction loss with additional regularization terms on the augmented samples:
\begin{align}
\mathcal{L}_{\text{reg}} = -\mathbb{E} \bigg[ \log D_{a}^{(y_i)}\big(\tilde{\mathbf{a}}_{i}\big) + \log D_{g}^{(s_i)}\big(\tilde{\mathbf{g}}_{i}\big) \bigg] \notag \\
-\frac{1}{|\mathcal{B}_{\text{AI}}^{(\text{aug})}|} \sum_{i \in \mathcal{B}_{\text{AI}}^{(\text{aug})}} \log D_{a}^{(y_i)}\big(\tilde{\mathbf{a}}^{(\text{aug})}_{i}\big),
\label{eq:p}
\end{align}
where $D^{(k)}(\cdot)$ denotes the predicted probability of class $k$ and $\mathcal{B}_{\text{AI}}^{(\text{aug})}$ is the set of augmented samples.

By disrupting residual cross-branch dependencies and injecting controlled semantic noise, the perturbation-enhanced regularization encourages structurally independent representations, ultimately enhancing robustness to unseen generator shifts.

\begin{table*}[t]
\centering
\fontsize{8.2}{9.5}\selectfont
\setlength{\tabcolsep}{6pt}
\renewcommand{\arraystretch}{1.15}
\begin{tabular}{lccccccc}
\toprule
\multirow{2}{*}{Categories} & \multicolumn{7}{c}{MAGE Corpus~\cite{li2024mage}} \\
\cmidrule(lr){2-8}
& FLAN-T5 & GPT & LLaMA & OPT & GLM & BigScience & EleutherAI \\
\midrule
\makecell[l]{\textit{LLMs} \\ \textit{included}}
& \makecell[c]{small/base/large\\xl/xxl} 
& \makecell[c]{davinci-002/003 \\ gpt-turbo-3.5} 
& \makecell[c]{6B/13B \\ 30B/65B} 
& \makecell[c]{2.7B/6.7B\\iml-1.3B/30B} 
& \makecell[c]{130B} 
& \makecell[c]{Bloom-7B} 
& \makecell[c]{GPT-J-6B \\GPT-NeoX-20B}  \\
\midrule
AI    & 9,000 & 13,800 & 9,000 & 9,000 & 9,100 & 8,900 & 9,000 \\
Human & 9,000 & 13,800 & 9,000 & 9,000 & 9,100 & 8,900 & 9,000 \\
\bottomrule
\end{tabular}
\caption{Number of AI-generated and human-written samples across datasets.
Balanced sampling is applied to categories with multiple LLMs so that each category contains a comparable number of samples, resulting in 20 representative models overall.}
\label{tab:datasets}
\end{table*}

\begin{table}[t]
\centering
\renewcommand{\arraystretch}{1.3}
\fontsize{8.5}{10}\selectfont
\begin{tabularx}{\linewidth}{@{}r X@{}}
\toprule
\multicolumn{2}{@{}l@{}}{\textbf{Algorithm 1: Structured Disentanglement Framework}} \\
\midrule
\multicolumn{2}{@{}l}{\makecell[l]{\textbf{Input:} encoder $E$, bottleneck encoders $E_a$, $E_g$,\\ discriminators $D_a$, $D_g$, weight factor $\beta$, training set $\mathcal{X}$}} \\
\multicolumn{2}{@{}l@{}}{\textbf{Output:} optimized encoders and discriminators} \\
\midrule
1  & \textbf{for} each training epoch \textbf{do} \\
2  & \quad \textbf{for} each mini-batch $\mathcal{B} = \{x_i, y_i, s_i\}_{i=1}^N$ \textbf{do} \\
3  & \quad\quad \textbf{Stage I: Encoding and Regularization} \\
4  & \quad\quad\quad $\mathbf{h}_i = E(x_i)$ \\
5  & \quad\quad\quad $\mathbf{e}_i^{(a)} \leftarrow \text{MLP}^{(a)}(\mathbf{h}_i)$, \quad $\mathbf{e}_i^{(g)} \leftarrow \text{MLP}^{(g)}(\mathbf{h}_i)$ \\
6  & \quad\quad\quad $q(\mathbf{a}_i|\mathbf{e}_i^{(a)})$, $q(\mathbf{g}_i|\mathbf{e}_i^{(g)}) \leftarrow$ Compute posteriors \\
6  & \quad\quad\quad $p(\mathbf{a})$, $p(\mathbf{g})\leftarrow$ Learnable priors \\
7  & \quad\quad\quad Compute DB loss $\mathcal{L}_{\text{DB}}$ via Eq.~\ref{eq:ib} \\
8  & \quad\quad\quad $\tilde{\mathbf{a}}_i$, $\tilde{\mathbf{g}}_i$, $\tilde{\mathbf{a}}^{(\text{aug})}_i \leftarrow$ Apply regularization \\
9  & \quad\quad\quad Compute discriminative loss $\mathcal{L}_{\text{reg}}$ via Eq.~\ref{eq:p} \\
10 & \quad\quad\quad Compute 1st-stage loss: 
     $\mathcal{L}_{\text{stage1}} = \beta \mathcal{L}_{\text{DB}} + \mathcal{L}_{\text{reg}}$ \\
11 & \quad\quad\quad Update $E$, $E_a$, $E_g$, $D_a$, $D_g$ w.r.t. $\mathcal{L}_{\text{stage1}}$ \\
12 & \quad\quad \textbf{Stage II: Discriminative-Guided Adaptation} \\
13 & \quad\quad\quad Freeze $D_a$, $D_g$ \\
14 & \quad\quad\quad $\mathbf{a}_i$, $\mathbf{g}_i \leftarrow$ pure features \\
15 & \quad\quad\quad Compute 2nd-stage loss 
     $\mathcal{L}_{\text{adapt}}$ via Eq.~\ref{eq:stage2} \\
16 & \quad\quad\quad Update $E_a$, $E_g$ w.r.t. $\mathcal{L}_{\text{adapt}}$ \\
\bottomrule
\end{tabularx}
\end{table}

\subsection{Discriminative-Guided Adaptation}
Despite the earlier regularization, residual overlaps between branches may persist due to shared cues. To further enforce disentanglement, we introduce a discriminative-guided adaptation stage.

Specifically, we freeze the parameters of the AI-detection discriminator $D_a$ and the generator-aware discriminator $D_g$, both pre-trained on perturbed representations to encode robust task semantics. 
For clean representations, each branch encoder is updated under two complementary constraints: 
(i) when passed into its own discriminator, the representation should be correctly classified, ensuring task alignment; 
(ii) when passed through a gradient reversal layer (GRL) into the opposite discriminator, the representation is encouraged to be misclassified, forcing the encoder to remove any information exploitable by the other branch.
Formally, the adaptation loss integrates both objectives:
\begin{align}
\mathcal{L}_{\text{adapt}} &= - \mathbb{E}  \bigg[\log D_{a}^{(y_i)}(\mathbf{a}_i) + \log D_{a}^{(y_i)}\big(\mathcal{G}(\mathbf{g}_i)\big) \nonumber \\
&  + \log D_{g}^{(s_{i})}(\mathbf{g}_i) +  \log D_{g}^{(s_i)}\big(\mathcal{G}(\mathbf{a}_i)\big)\bigg],
\label{eq:stage2}
\end{align}
where $\mathcal{G}(\cdot)$ denotes the GRL operation. 
This setup ensures each encoder fits its own semantics while eliminating cross-branch cues. By refining encoders under frozen discriminators with clean inputs, this stage yields more invariant, disentangled, and generalizable representations. The complete procedure, including loss definitions and optimization steps, is summarized in Algorithm 1.

\begin{table*}[t]
\fontsize{8}{9.5}\selectfont
\setlength\tabcolsep{7.5pt}
\renewcommand{\arraystretch}{1.2}
\centering
\begin{tabular}{lccccccc}
\toprule
& \multicolumn{7}{c}{Accuracy (\%) / $F_1$-Measure} \\
\cmidrule(lr){2-8}
LLM Categories & FLAN-T5 & GPT & LLaMA & OPT & GLM & BigScience & EleutherAI \\
\midrule
\multicolumn{8}{l}{\textit{Zero-shot methods}} \\
$\text{GLTR}_{\text{ACL'19}}$ & 56.9 / 52.4 & 71.8 / 71.5  & 75.1 / 75.8 & 75.1 / 77.3 & 77.8 / 78.2 & 86.6 / 86.4 & 80.2 / 80.0 \\
$\text{Fast-DetectGPT}_{\text{ICLR'24}}$ & 57.3 / 74.9 & 70.0 / 82.5 & 73.4 / 79.8 & 74.6 / 76.1 & 69.9 / 72.8 & 79.3 / 84.2 & 77.0 / 82.9 \\
$\text{MCP}_{\text{ACL'25}}$ & 58.6 / 76.3 &  72.1 / 84.4 & 75.9 / 81.5 & 77.4 / 78.3 & 74.7 / 75.2 & 86.2 / 86.5 & 83.7 / 84.4 \\
\midrule
\multicolumn{8}{l}{\textit{Training-based methods}} \\
$\text{BERT}_{\text{NAACL'19}}$ & 60.9 / 52.6 & 78.6 / 77.5  & 83.1 / 85.4 & 82.1 / 84.2 & 92.4 / 92.8 & 96.4 / 95.7 & 96.7 / 95.9  \\
$\text{IT:CLUST}_{\text{ACL’22}}$ & 58.7 / 57.5 & 63.1 / 61.0 & 71.3 / 70.4 & 72.1 / 74.7 & 75.7 / 77.1  & 85.4 / 87.9 & 82.5 / 81.8 \\
$\text{PECOLA}_{\text{ACL'24}}$ & 65.1 / 65.5 & 81.1 / 80.3 & 84.6 / 84.5 & 83.5 / 83.4 & 78.3 / 77.8 & 87.6 / 86.7 & 83.9 / 84.5 \\
$\text{SCRN}_{\text{ACL'24}}$ & 59.6 / 58.7 & 69.0 / 67.8 & 75.0 / 73.3 & 81.3 / 79.2 & 71.8 / 67.7 & 87.8 / 87.1 & 90.6 / 90.0 \\

\rowcolor{gray!7}\textbf{Ours} & \textbf{68.9} / \textbf{64.1} & \textbf{82.7} / \textbf{82.6} & \textbf{88.0} / \textbf{88.5} & \textbf{89.1} / \textbf{88.3} & \textbf{94.1} / \textbf{93.9} & \textbf{96.7} / \textbf{96.4} & \textbf{97.2} / \textbf{97.8} \\
\bottomrule
\end{tabular}
\caption{Cross-generator generalization results on seven LLM categories from the MAGE benchmark under the leave-one-generator-out protocol. Zero-shot and training-based methods are compared against our approach, which yields consistent gains across all held-out generators.}
\label{tab:main}
\end{table*}

\section{Experiments}
\subsection{Experimental Details}
\label{Experimental_Details}
\textbf{Datasets.} 
We adopt MAGE benchmark~\cite{li2024mage} to comprehensively evaluate cross-generator generalization in AIGT detection. 
The original dataset consists of 27 LLMs grouped into 7 categories.
To prioritize generalization over redundancy, we remove highly similar variants of the same categories, thereby obtaining a subset of 20 representative models.
For category with a single model, the entire set is retained, whereas for those with multiple sub-models, we apply balanced random sampling to ensure comparable sizes.
As summarized in Table~\ref{tab:datasets}, the final dataset spans diverse generator families (e.g., GPT, LLaMA, FLAN-T5), thereby providing a broad and challenging testbed for cross-generator evaluation.


\textbf{Competitive Methods.} 
To evaluate generalization on unseen generators, we compare with SOTA methods in two categories: training-based and zero-shot. 
Training-based methods include PECOLA~\cite{liu2024does} and SCRN~\cite{huang2024ai}, which add input noise or regularization to improve stability, and IT:CLUST~\cite{shnarch2022cluster}, which exploits clustering for intermediate supervision and domain adaptation. 
Zero-shot methods include Fast-DetectGPT~\cite{bao2024fast} and GLTR~\cite{gehrmann2019gltr}, relying on pretrained LM statistics, and MCP~\cite{zhu2025reliably}, which uses conformal prediction to calibrate thresholds and control the false positive rate. 
We also report BERT~\cite{devlin2019bert} as an encoder-only baseline.
These methods provide baselines for evaluating robustness and generalization.

\textbf{Implementation Details.} We use the Adam optimizer with a learning rate of $2 \times 10^{-5}$ and a batch size of 16 across all experiments.
During the DB-based encoding stage, each latent variable is sampled 5 times per instance to estimate expectations.
Both the AIGT and generator discrimination heads employ a dropout rate of 0.5 to mitigate overfitting.
The loss balancing coefficient $\beta$, which controls the trade-off between the DB regularization and the discriminative objective in both stages, is set to $5\times 10^{-6}$.
All experiments are conducted on two NVIDIA GeForce RTX 3090 GPUs.

\subsection{Overall Performance}
\subsubsection{Main Results on Cross-Generator Generalization}
We evaluate our method under the leave-one-generator-out setting, training on 6 generator categories and testing on the held-out one. 
As shown in Table~\ref{tab:main}, accuracies vary widely from 60\% to 98\%, reflecting substantial variation in cross-generator difficulty.
Across this range, our model consistently outperforms all competitive methods, achieving up to 24.2\% accuracy improvement over Fast-DetectGPT on unseen GLM set.
The gains are particularly notable under large distributional shifts.
For instance, on the challenging unseen FLAN-T5 our approach improves accuracy by 8.0\% over BERT, while on the high-performing EleutherAI, it yields 0.5\%. 
These results highlight that our disentanglement design enables robust cross-generator generalization, particularly under large distributional shifts between train/test generators.

\begin{table}[t]
\centering
\fontsize{8}{9.5}\selectfont
\setlength\tabcolsep{5pt}
\renewcommand{\arraystretch}{1.15}
\begin{tabular}{lcccc}
\toprule
\multirow{2}{*}{Model} & \multicolumn{4}{c}{Accuracy (\%) / $F_{1}$-Measure} \\
\cmidrule(lr){2-5}
& $N=2$ & $N=3$ & $N=4$ & $N=5$ \\
\midrule
\multicolumn{5}{c}{OPT Test} \\
\hline
BERT$_{\text{NAACL'19}}$            & 72.8/72.2 & 73.1/69.9 & 76.7/74.9 & 78.1/76.1\\
$\text{IT:CLUST}_{\text{ACL’22}}$ & 69.4/67.7 & 68.4/69.8 & 70.1/68.1 & 68.4/68.1 \\
PECOLA$_\text{ACL'24}$ & 70.1/68.1 & \textbf{76.8/75.5} & 80.5/80.0 & 81.1/80.9 \\
SCRN$_\text{ACL'24}$   &69.2/68.3  & 70.7/67.4 & 72.5/69.9 & 76.1/75.9 \\
\rowcolor{gray!10}
\textbf{Ours}       & \textbf{73.1/74.7} & 73.9/75.1 & \textbf{80.5/80.4} & \textbf{83.2/82.3} \\
\midrule
\multicolumn{5}{c}{FLAN-T5 Test} \\
\hline
BERT$_{\text{NAACL'19}}$            & 51.2/50.2 & 52.9/51.1 & 55.0/53.9 & 58.1/57.8 \\
$\text{IT:CLUST}_{\text{ACL’22}}$ & 53.1/53.7 & 54.4/53.1 & 54.2/53.0 & 53.6/52.2 \\
PECOLA$_\text{ACL'24}$ & \textbf{56.7}/\textbf{56.3} & 54.2/50.2 & 57.4/57.1 & 61.3/60.1 \\
SCRN$_\text{ACL'24}$   & 53.6/52.1 & 54.1/53.2 & 56.8/55.9 & 57.8/57.4 \\
\rowcolor{gray!10}
\textbf{Ours}       & 54.7/52.6 & \textbf{57.0/55.1} & \textbf{61.1/58.4} & \textbf{64.5/62.4}  \\
\bottomrule
\end{tabular}
\caption{Cross-generator generalization on OPT and FLAN-T5. 
Models are trained with $N$ categories under a fixed training size of 12,000 samples and evaluated on an unseen LLM category.}
\label{tab:diversity}
\end{table}

\begin{figure*}[t]
    \centering
    \includegraphics[scale=.38]{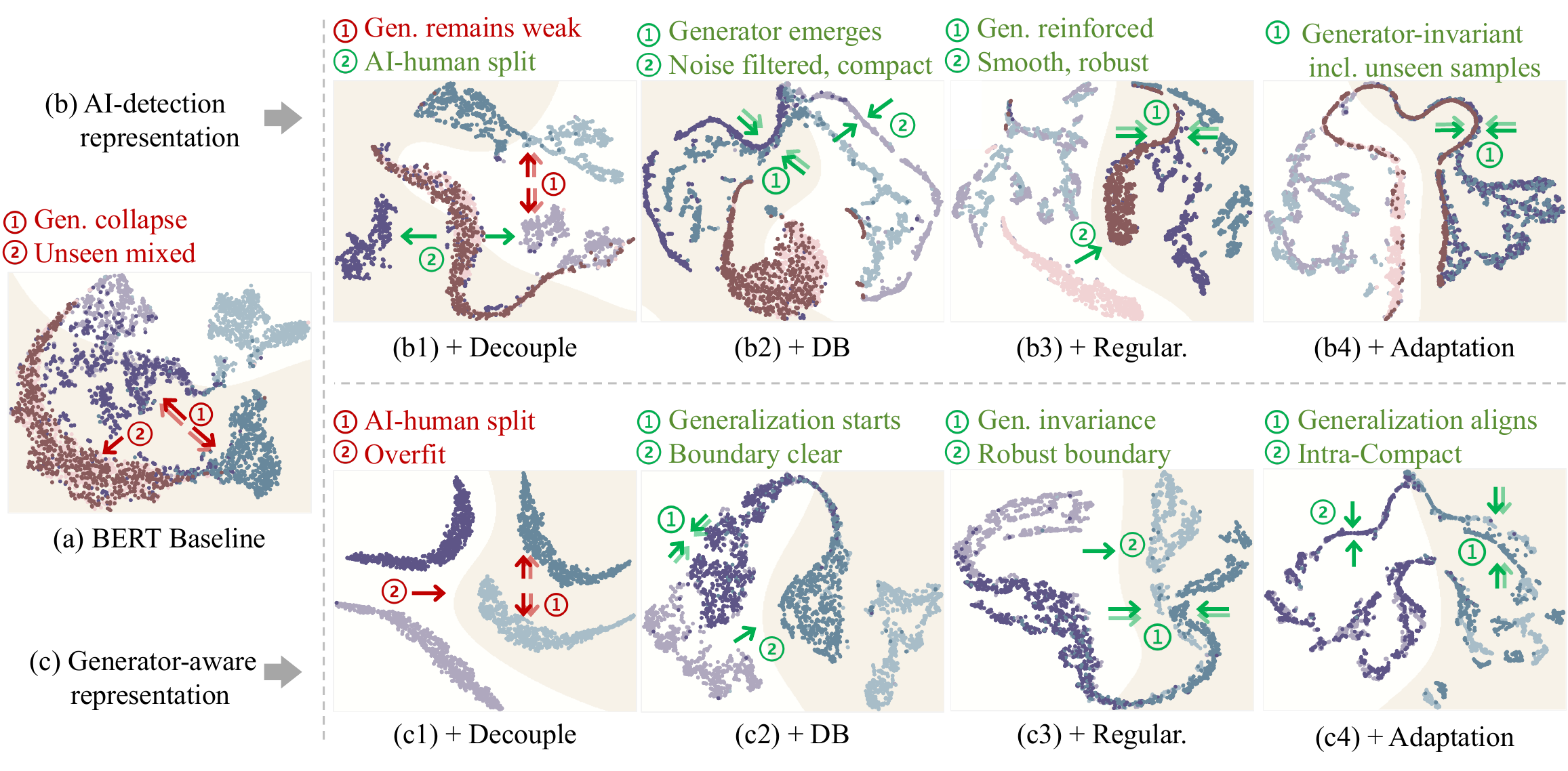} 
     \caption{Progressive t-SNE visualization of disentangled feature spaces. 
     The BERT baseline (a) collapses, with unseen samples (red) mixed into training clusters (blue/purple). 
     After decoupling, the AI-detection (b) and generator-aware (c) branches begin to separate (b1/c1), but generalization remains unstable and generator-sensitive. 
     With DB encoding (b2/c2), semantic noise is reduced, yielding more compact clusters. 
    Cross-view regularization (b3/c3) smooths boundaries and enhances robustness. 
     Finally, discriminator-guided adaptation (b4/c4) yields clear human–AI separation in the detection branch and generator-invariant alignment in the auxiliary branch.}
    \label{fig:generalization}
\end{figure*}

\subsubsection{Effect of Training Generator Diversity}
To examine the effect of training diversity on cross-generator generalization, we hold out OPT and FLAN-T5 as test categories while training on subsets of the remaining 5 families. 
For each setting, we fix the training size at 12,000 instances and vary the number of training generators $N \in \{2,3,4,5\}$, distributing samples evenly across the selected categories to control for data volume. 
As shown in Table~\ref{tab:diversity}, our method consistently benefits from increased diversity and demonstrates stronger scalability than recent alternatives. 
In the unseen OPT detection scenario, under low-diversity training ($N=2$), the margin over the strongest method is minimal (0.3\% over BERT), but under high-diversity training ($N=5$) it expands substantially (5.1\% over BERT). 
These findings confirm that training diversity, rather than training volume, is the key driver for learning invariant features and achieving robust cross-generator generalization.

\subsection{Generalization via Progressive Cumulative Design}
We visualize the evolution of latent features as modules are progressively added.
Models are trained on GPT and BigScience categories and evaluated on the unseen FLAN-T5 one, with t-SNE projections shown in Figure~\ref{fig:generalization}.
The BERT encoder produces entangled and poorly aligned representations, while each additional module: decoupling, DB encoding, cross-view regularization, and adaptation incrementally enhances compactness, robustness, and ultimately cross-generator generalization.

\subsection{Robustness under Adversarial Attacks}
We evaluate robustness under two types of adversarial attacks on the unseen OPT category, including word-level perturbations and model-aware adversarial attacks, as summarized in Table~\ref{tab:robustness_combined}. Under word-level perturbations (deletion, insertion, swap, and replacement), our method consistently outperforms BERT and SCRN, exhibiting a smaller performance drop from clean inputs, especially under deletion. Furthermore, measured by attack success rate (ASR), our approach is significantly more resilient to model-aware attacks (PWW~\cite{ren2019generating}, TextFooler~\cite{jin2020bert}, and Deep-Word-Bug~\cite{gao2018black}) compared to the highly vulnerable BERT baseline. These findings confirm that disentangled, generator-invariant representations effectively improve model robustness against diverse adversarial manipulations under cross-generator shifts.
\begin{table}[t]
\centering
\fontsize{8.5}{10}\selectfont
\begin{tabular}{lccc}
\toprule
\multicolumn{4}{c}{\textbf{Word-level Perturbations}} \\
\midrule
Metric & \multicolumn{3}{c}{Accuracy (\%) / $F_1$-Measure}  \\
\cmidrule(lr){2-4} 
Model & BERT & SCRN & Ours \\
\midrule
Original & 82.1 / 84.2 & 81.3 / 79.2 & \textbf{89.1} / \textbf{88.3} \\
\midrule
Delete   & 68.8 / 72.6 & 67.6 / 64.9 & \textbf{80.0} / \textbf{79.9} \\
Swap     & \textbf{66.9} / 64.2 & 62.6 / 59.9 & 66.5 / \textbf{69.9} \\
Insert   & 66.9 / 65.1 & 62.6 / 60.9 & \textbf{67.5} / \textbf{70.6} \\
Replace  & 66.4 / 65.3 & 63.5 / 66.9 & \textbf{74.5} / \textbf{75.6} \\
\midrule
Average  & 67.3 / 66.8  & 64.1 / 63.2 & \textbf{72.1} / \textbf{74.0} \\
\midrule
\multicolumn{4}{c}{\textbf{Adversarial Attacks (ASR $\downarrow$)}} \\
\midrule
Model & BERT & SCRN & Ours \\
\midrule
PWWS        & 30.7 & 27.4 & \textbf{27.2} \\
TextFooler        & 61.5 & 58.7 & \textbf{54.5} \\
Deep-Word-Bug    & 46.1 & 30.9 & \textbf{22.4} \\
\bottomrule
\end{tabular}
\caption{Robustness comparison under two types of attacks on the unseen OPT category.
Word-level perturbation are evaluated using accuracy and $F_1$ scores, while adversarial attacks are reported by attack success rate (ASR).
Lower ASR indicates stronger robustness.}
\label{tab:robustness_combined}
\end{table}

\subsection{Analysis of Core Modules}
To gain deeper insights into generalization, we study how individual modules influence the representation space along three dimensions: disentanglement, robustness, and compactness.

\begin{figure}[t]
    \centering
    \includegraphics[width=\columnwidth]{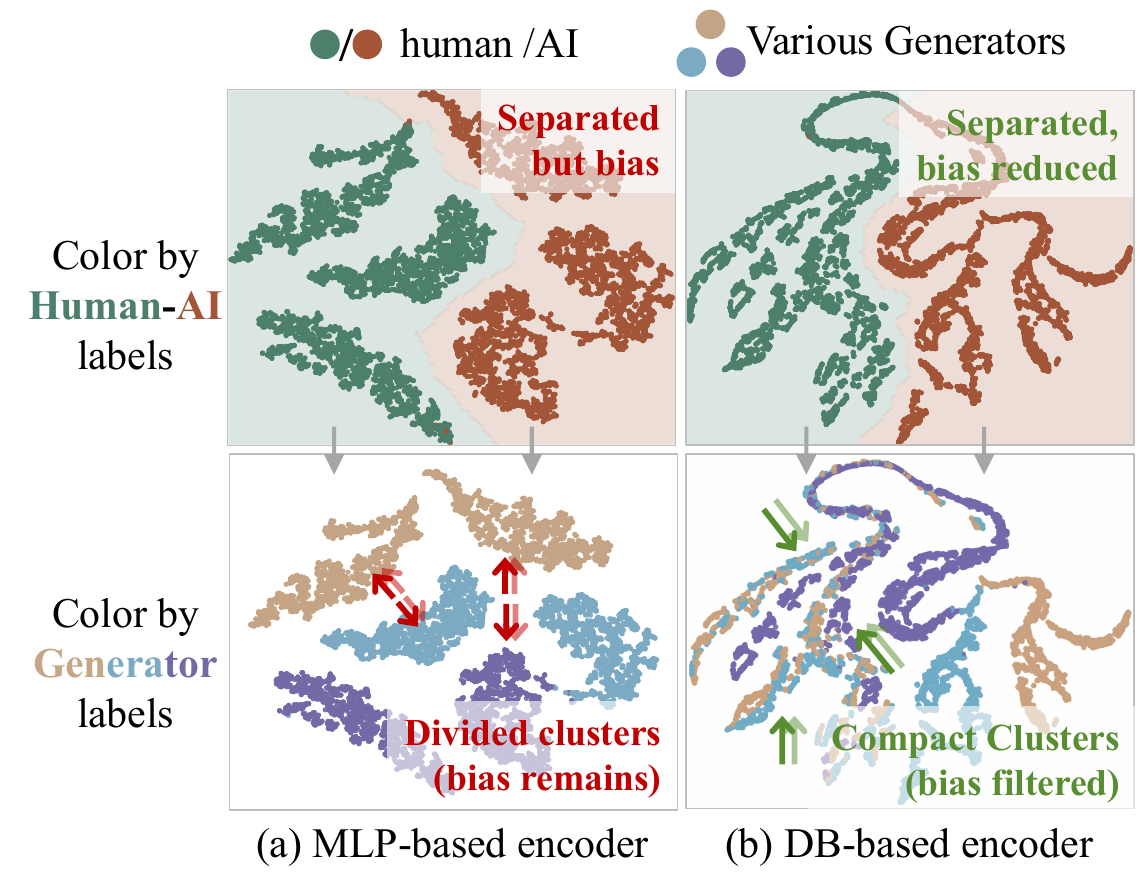} 
     \caption{T-SNE visualizations of AI-detection features $\mathbf{a}$ from MLP and DB encoders.
     With human-AI coloring (top), the encoders separate authenticity.
     With generator coloring (bottom), MLP features split into distinct generator clusters, while DB suppresses generator bias, yielding compact, generator-invariant representations.}
    \label{fig:ib}
\end{figure}


\subsubsection{Filtering Generator Bias via Bottlenecked Encoding}
To assess the contribution of the DB encoding module, we compare the AI-detection representations $\mathbf{a}$ learned by a conventional MLP encoder and our DB encoder, both trained on GPT, BigScience, and FLAN-T5 sets.
As shown in Figure~\ref{fig:ib}, both models successfully separate human from AI texts, but the MLP encoder exhibits clear generator bias with divided clusters.
In contrast, our DB encoder mitigates such bias, yielding compact and generator-invariant features that support invariant feature learning and substantially enhance cross-generator generalization.
Additionally, the complementary analysis of the representations learned by the generator-aware branch $\mathbf{g}$ is detailed in Appendix~\ref{sec:appendix_bottleneck}.

\begin{figure}[t]
    \centering
    \includegraphics[scale=.35]{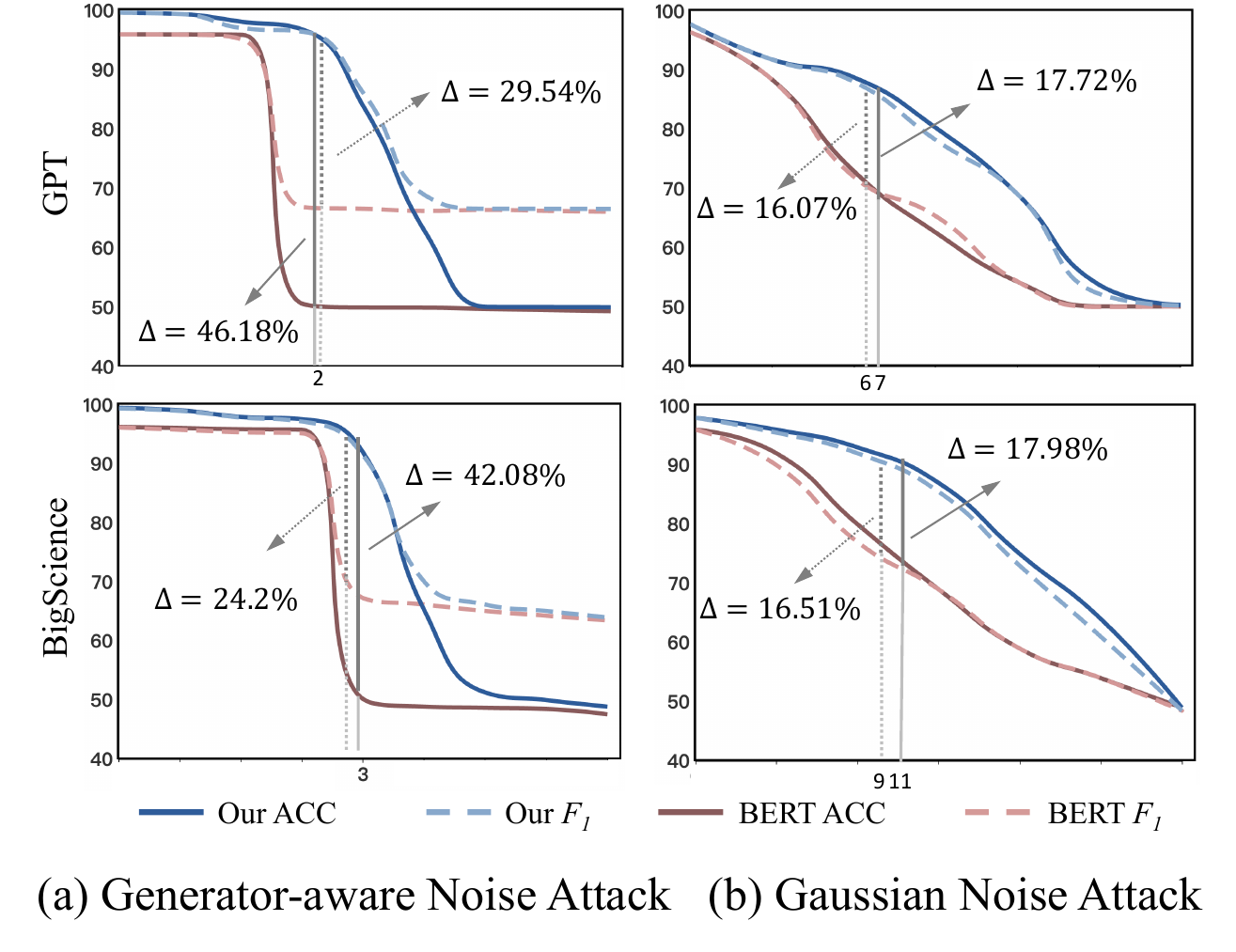} 
     \caption{Robustness under generator-aware and Gaussian perturbations. Perturbation-based regularization enhances our method’s resilience, maintaining superior accuracy over the BERT baseline in both structured and random noise settings.}
    \label{fig:perturb}
\end{figure}

\subsubsection{Enhancing Robustness via Cross-View Regularization}
To evaluate the robustness benefits of the cross-view regularization module, we inject two types of noise into the AI-detection features: (1) generator-aware perturbations simulating stylistic interference, and (2) Gaussian noise introducing unstructured variation. 
We compare the performance of model variants under both noise conditions.
As shown in Figure~\ref{fig:perturb}, our model consistently maintains significantly higher classification accuracy, with improvements of up to 46.18\% (on GPT) under generator-style perturbations and 17.98\% (on BigScience) under Gaussian noise. 
These results confirm that cross-view regularization enhances robustness by reducing sensitivity to both structured and unstructured perturbations, thereby improving generalization under distributional shifts.

\begin{figure}[t]
    \centering
    \includegraphics[scale=.4]{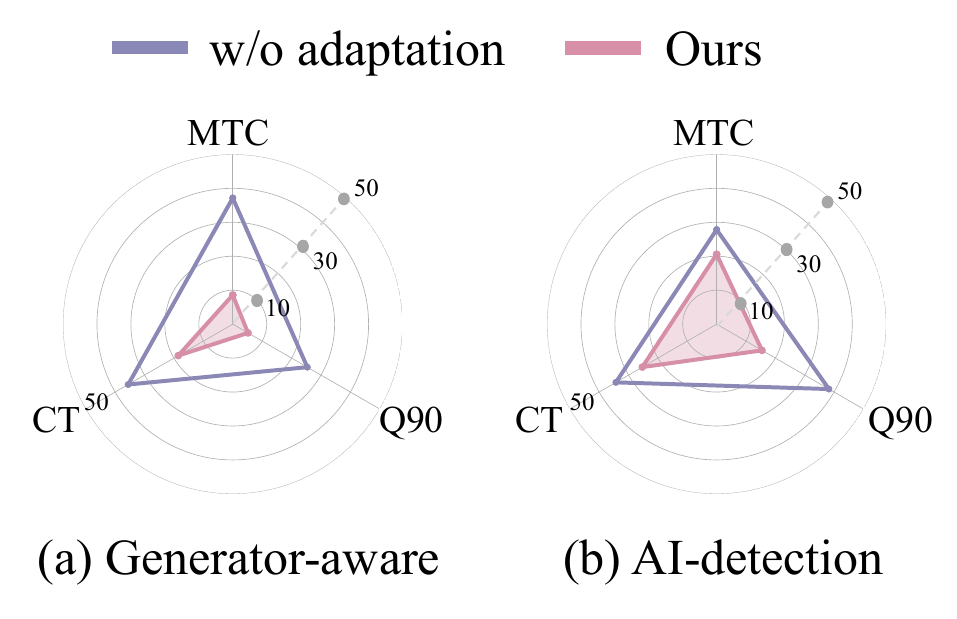} 
     \caption{Intra-class compactness with vs. without adaptation. Adaptation consistently improves feature compactness across tasks, yielding lower dispersion metrics and more task-aligned representations.}
    \label{fig:refine}
\end{figure}

\subsubsection{Improving Compactness via Adaptation}
To evaluate the impact of the discriminator-guided adaptation stage, we analyze intra-class compactness of the learned features. 
Specifically, we compute three metrics: MeanToCenter, Covariance Trace, and 90th Percentile Pairwise Distance, on AI-detection and generator-aware representations extracted from GPT category. 
As shown in Figure~\ref{fig:refine}, models trained with the adaptation module consistently exhibit lower values across all metrics, indicating tighter clustering and reduced semantic noise. 
These findings suggest that the adaptation module facilitates alignment, leading to more compact and discriminative feature structures, thereby enhance model generalization.

\begin{table}[t]
\fontsize{8.5}{10}\selectfont
\setlength\tabcolsep{2pt}
\renewcommand{\arraystretch}{1.2}
\centering
\begin{tabular}{ccccccc}
\toprule
\multicolumn{5}{c}{Components} &\multicolumn{2}{c}{Accuracy (\%) / $F_1$-Measure}  \\
\cmidrule(r){1-5} \cmidrule(lr){6-7} 
BERT & Disen. & DB & Regular. & Adapt. & FLAN-T5 & OPT \\
\midrule
\checkmark &       &       &       &        &  60.9 / 52.6 & 82.1 / 84.2 \\
\checkmark & \checkmark &       &       &        & 62.0 / 53.3 & 83.7 / 83.5 \\
\checkmark & \checkmark & \checkmark &       &        & 65.2 / 60.2 & 87.1 / 86.0 \\
\checkmark & \checkmark & \checkmark & \checkmark &   & 66.8 / 63.8 & 88.3 / 87.4 \\
\checkmark & \checkmark & \checkmark & \checkmark & \checkmark & \textbf{68.9} / \textbf{64.1} & \textbf{89.1} / \textbf{88.3} \\
\bottomrule
\end{tabular}
\caption{Incremental ablation results on two unseen LLM categories.}
\label{tab:ablation}
\end{table}


\subsection{Ablation Study}
We conduct an incremental ablation study on the FLAN-T5 and OPT as the held-out evaluation category set (see Table~\ref{tab:ablation}).
For each test case, the model is trained on the remaining 6 categories and evaluated on the selected one to measure generalization.
Adding DB-based encoding provides the largest gain by filtering generator noise and clarifying semantics.
Cross-view regularization improves robustness by stabilizing representations under perturbations, and discriminator-guided adaptation further refines decision boundaries for compact, task-aligned features.
Overall, each module contributes additive improvements, with DB-based encoding as the most influential.

\section{Conclusion}
We propose a structured disentanglement framework for detecting AIGT content from unseen generators, tackling distribution shifts and stylistic artifacts. By combining DB-based encoding, cross-view regularization, and discriminator-guided adaptation, our method suppresses generator-aware noise and improves generalization. Extensive experiments demonstrate state-of-the-art cross-generator performance, highlighting the scalability of disentanglement for AIGT detection.

\section*{Limitations}
While our framework achieves strong cross-generator generalization, several limitations remain. 
One concern lies in interpretability: although feature visualizations provide some intuition, the specific linguistic or structural cues driving detection decisions are not yet fully understood. 
Another issue is sensitivity to hyperparameter choices, which may require careful tuning to maintain stable performance across settings. 
In addition, robustness against adaptive or adversarial attacks has not been systematically examined, and strengthening this aspect would be essential for reliable deployment in real-world applications.

\section*{Acknowledgements}
This work was supported partly by the National Natural Science Foundation of China (62402073, 62403093, U22A2096 and 62221005) and the Science and Technology Research Program of Chongqing Municipal Education Commission (KJQN202300619, KJQN202300606 and KJQN202400650).


\bibliography{custom}


\appendix

\section{Appendix}
\begin{table*}[t]
\fontsize{8}{9.5}\selectfont
\setlength\tabcolsep{7.5pt}
\renewcommand{\arraystretch}{1.2}
\centering
\begin{tabular}{lccccccc}
\toprule
& \multicolumn{7}{c}{Accuracy (\%) / $F_1$-Measure} \\
\cmidrule(lr){2-8}
LLM Categories  & FLAN-T5 & GPT & LLaMA & OPT & GLM & BigScience & EleutherAI \\
\midrule
$\text{BERT}_{\text{NAACL'19}}$ 
& 70.5 / 66.5 
& 83.9 / 82.7 
& 87.4 / 88.5 
& 86.5 / 87.3 
& 93.1 / 93.9 
& 98.1 / \textbf{97.3} 
& 98.0 / 97.1 \\
$\text{PECOLA}_{\text{ACL'24}}$ 
& 75.5 / 75.8
& 85.6 / 85.1 
& 88.9 / 88.7 
& 88.5 / 88.3 
& 80.9 / 80.2 
& 90.4 / 89.6 
& 86.7 / 87.4  \\
$\text{SCRN}_{\text{ACL'24}}$ 
& 69.1 / 68.5
& 74.6 / 73.7 
& 79.7 / 78.3 
& 85.6 / 83.9 
& 73.4 / 70.2 
& 89.5 / 89.1 
& 92.1 / 91.8  \\
\rowcolor{gray!7}\textbf{Ours}
& \textbf{77.4} / \textbf{77.1}
& \textbf{86.9} / \textbf{87.4}
& \textbf{91.7} / \textbf{90.1}
& \textbf{93.8} / \textbf{92.4}
& \textbf{95.7} / \textbf{94.6}
& \textbf{98.7} / 96.1
& \textbf{98.3} / \textbf{98.9} \\
\bottomrule
\end{tabular}
\caption{Performance comparison across seven LLM generator families on the full original MEGA dataset.}
\label{tab:in-generation}
\end{table*}

\subsection{Data Preprocessing}
\label{Data_Preprocessing}
We adopt the pre-processed dataset from the MAGE benchmark~\cite{li2024mage}, which provides both generator categories and domain labels. 
To ensure consistency with the original distribution, we perform balanced sampling for each category by aligning with the proportions of both generator and domain labels. 
This design controls for domain effects and encourages the model to learn generalization across generators rather than being confounded by domain-specific biases. 
For clarity and to avoid redundancy, we exclude several highly similar models, including T0-11B, T0-3B, OPT-1.3B, OPT-125M, OPT-13B, OPT-30B, and OPT-350M.

To validate the effectiveness and robustness of this data preprocessing strategy, we conduct additional evaluations on both the original MAGE dataset and more recent large-scale generators.

\begin{table}[t]
\centering
\fontsize{8.5}{10}\selectfont
\begin{tabular}{lccc}
\toprule
Metric & \multicolumn{3}{c}{Accuracy (\%) / $F_1$-Measure} \\
\cmidrule(lr){2-4} 
Model& BERT & SCRN & Ours \\
\midrule
GPT-4 & 52.8 / 51.0 & 51.6 / 50.9 & \textbf{61.8} / \textbf{65.4} \\
GPT-5  & 61.0 / 61.6 & 53.8 / 52.2 & \textbf{68.7} / \textbf{70.1} \\
Claude-sonnet-4-5 & 51.6 / 56.6 & 52.9 / 52.1 & \textbf{57.1} / \textbf{60.5} \\
DeepSeek-R1  & 63.3 / 69.7 & 55.4 / 53.2 & \textbf{71.0} / \textbf{69.8} \\
\bottomrule
\end{tabular}
\caption{Performance comparison among BERT, SCRN, and our method on the GPT-4 OOD split in MAGE (AI: 762, Human: 800), along with $\sim$800 manually collected samples from GPT-5, Claude, and DeepSeek.}
\label{tab:OOD_large_models}
\end{table}

\begingroup
\renewcommand\cellgape{}              
\setlength{\aboverulesep}{0pt}        
\setlength{\belowrulesep}{0pt}
\renewcommand{\arraystretch}{0.8}     
\begin{table}[t]
\centering
\fontsize{8.5}{10}\selectfont
\begin{tabularx}{\columnwidth}{lXX}
\toprule
\multicolumn{1}{c}{Section} & 
\multicolumn{1}{c}{Train} & 
\multicolumn{1}{c}{Test} \\
\toprule
\arrayrulecolor{gray!40}
\makecell[l]{4.2.1 Main Results \\ (Table~\ref{tab:main})}     
    & \multicolumn{2}{c}{\makecell{Leave-One-Generator-Out}} \\
\midrule
\makecell[l]{4.2.2 Diversity \\ (Table~\ref{tab:diversity})}        
    & \makecell[l]{BigScience \\ GPT \\ LLaMA \\ GLM \\ EleutherAI} 
    & \makecell{OPT \\ FLAN-T5} \\
\midrule
\makecell[l]{4.3 Progress Gen. \\ (Fig.~\ref{fig:generalization})}    
    & \makecell[l]{BigScience \\ GPT} 
    & \makecell[l]{FLAN-T5} \\
\midrule
\makecell[l]{4.4.1 Bias Filter (Fig.~\ref{fig:ib}) \\ A.5 Bias Filter (Fig.~\ref{fig:ib_option})} 
    & \makecell[l]{BigScience \\ GPT \\ FLAN-T5} 
    & \makecell[c]{-} \\
\midrule
\makecell[l]{4.4.2 Robustness \\ (Fig.~\ref{fig:perturb})}      
    & \makecell[l]{BigScience\\GPT} 
    & \makecell[c]{-}\\
\midrule
\makecell[l]{4.4.3 Compactness \\ (Fig.~\ref{fig:refine})}     
    & \makecell[l]{GPT} 
    & \makecell[c]{-}\\
\midrule
\makecell[l]{4.5 Ablation Study \\ (Table~\ref{tab:ablation})}     
    & \makecell[l]{BigScience \\ GPT \\ LLaMA \\ GLM \\ EleutherAI}  
    & \makecell{FLAN-T5 \\ OPT} \\
\midrule
\makecell[l]{A.1.1 Eval. on Original Dataset \\ (Table~\ref{tab:in-generation})} 
    & \multicolumn{2}{c}{Leave-One-Generator-Out} \\
\midrule
\makecell[l]{A.1.2 Eval. on OOD Dataset \\ (Table~\ref{tab:OOD_large_models})} 
    & \makecell{MAGE} 
    & \makecell{$\text{OOD}_i$} \\
\midrule
\makecell[l]{A.3 Adver. Attacks \\ (Table~\ref{tab:robustness_combined})} 
    & \makecell[l]{$\bigcup_{i \neq j} \text{Test}_i$} 
    & \makecell{OPT} \\
\midrule
\makecell[l]{A.4 Generalization \\ (Fig.~\ref{fig:gen_option})}    
    & \shortstack[l]{$\bigcup_{i \neq j} \text{Test}_i$}
    & \makecell{OPT \\ FLAN-T5 \\ LLaMA} \\
\arrayrulecolor{black}\bottomrule
\end{tabularx}
\caption{Overview of training and testing setups across different experiments.}
\label{tab:exp-setup}
\end{table}
\endgroup

\subsubsection{Evaluation on Original MEGA dataset}
To further verify that our results are not an artifact of re-sampling or generator filtering, we also report results on the full original MAGE splits under the standard 6-vs-1 evaluation protocol (Table~\ref{tab:in-generation}).
Consistent performance trends are observed across both settings.
Due to the inclusion of additional highly similar generators in the original split, certain categories achieve slightly higher absolute scores.
Nevertheless, our method consistently outperforms the baselines, demonstrating its general robustness applicability across different data distributions and generator configurations.

\subsubsection{Evaluation on Latest SOTA Generators}
We further evaluate unseen-generator generalization on an out-of-distribution (OOD) split involving recent large-scale models, including the GPT-4 OOD split from MAGE and manually collected samples from GPT-5, Claude-sonnet-4-5, and DeepSeek-R1, unseen during training (Table~\ref{tab:OOD_large_models}).
As shown in the table, both BERT and SCRN exhibit noticeable performance degradation on these advanced generators.
In contrast, our method consistently achieves higher accuracy and $F_1$ scores across all models, indicating more stable generalization under increasing model scale and generation quality.
This suggests that disentangled representations capture generator-invariant cues that remain effective for emerging LLMs.

\subsection{Overview of Training and Testing Setups}
Table \ref{tab:exp-setup} summarizes the training and testing setups used across all experiments. 
For each section, we specify the generator categories included in the training set and those held out for testing.  
In particular, Section 4.2.1 adopts a leave-one-generator-out protocol, while subsequent experiments examine generalization under varying generator diversity, progressive training scenarios, robustness to bias and adversarial perturbations, and ablation analyses.  
This overview facilitates a clear understanding of how different experimental objectives are mapped to corresponding train–test splits, thereby providing a unified and consistent evaluation protocol across all settings.

\subsection{Generalization Gains from Feature Disentanglement}
\begin{figure}[t]
    \centering
    \includegraphics[width=\columnwidth]{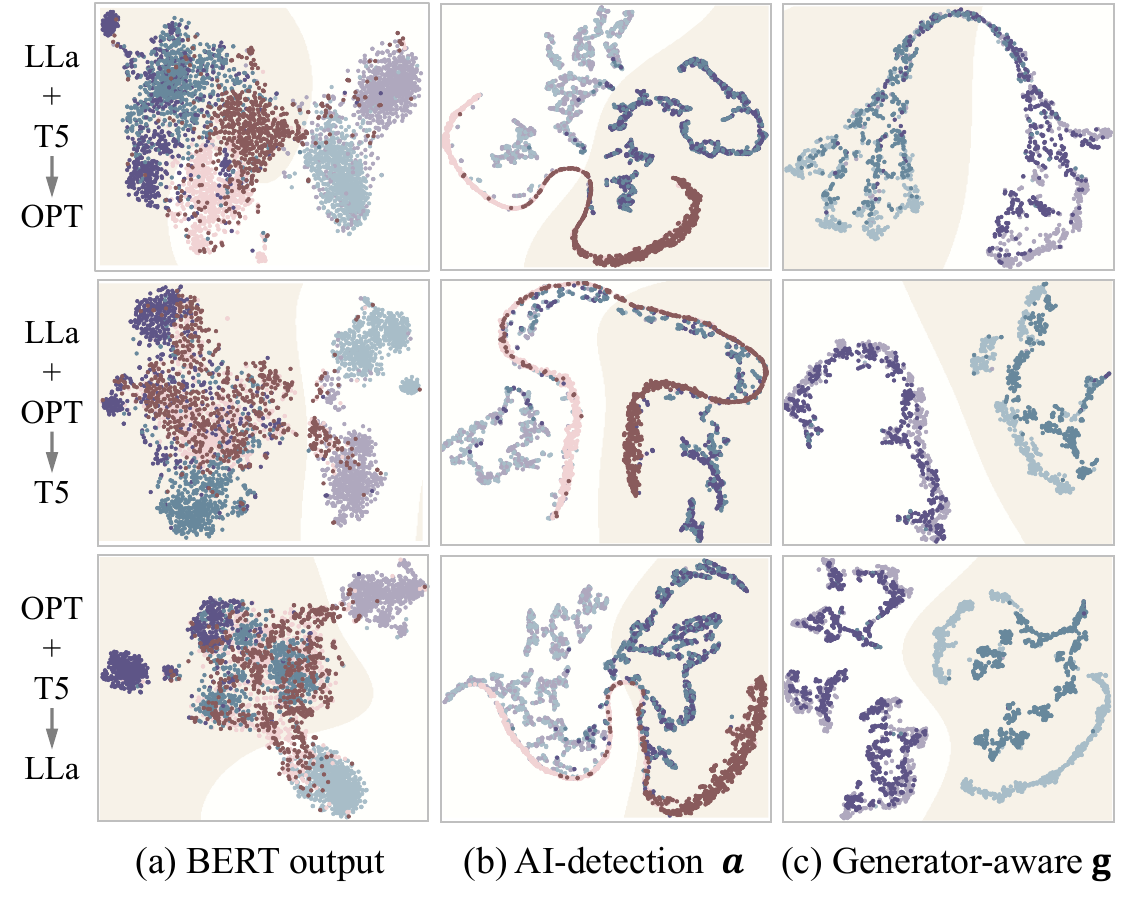} 
     \caption{T-SNE visualizations across LLaMA, OPT, and FLAN-T5. Training (blue/purple) and held-out (red) generators are shown, with dark/light shades for AI/human texts. (a) Baseline BERT entangles features on unseen generators. Conversely, our framework yields (b) tighter human–AI separation with mitigated generator bias in the AI-detection branch, and (c) explicit isolation of generator-specific features in the generator-aware branch.}
    \label{fig:gen_option}
\end{figure}

To assess cross-generator generalization, we conduct a leave-one-generator-out analysis on three MAGE categories (LLaMA, OPT, FLAN-T5). 
In each run, models are trained on two generators and tested on the remaining unseen one, with representation distributions visualized via t-SNE (Figure~\ref{fig:gen_option}). 
As shown in panel (a), BERT features separate human and AI samples on seen generators but fail to generalize to the unseen one, where generator entanglement persists. 
With disentanglement, the AI-detection branch (panel b) enlarges the human–AI margin while suppressing generator bias, and the generator-aware branch (panel c) captures generator distinctions explicitly. 
These results demonstrate that disentanglement isolates task-relevant semantics from generator artifacts, leading to improved cross-generator generalization.

\subsection{Further Analysis on Bottlenecked Encoding}
\label{sec:appendix_bottleneck}
\begin{figure}[t]
    \centering
    \includegraphics[width=\columnwidth]{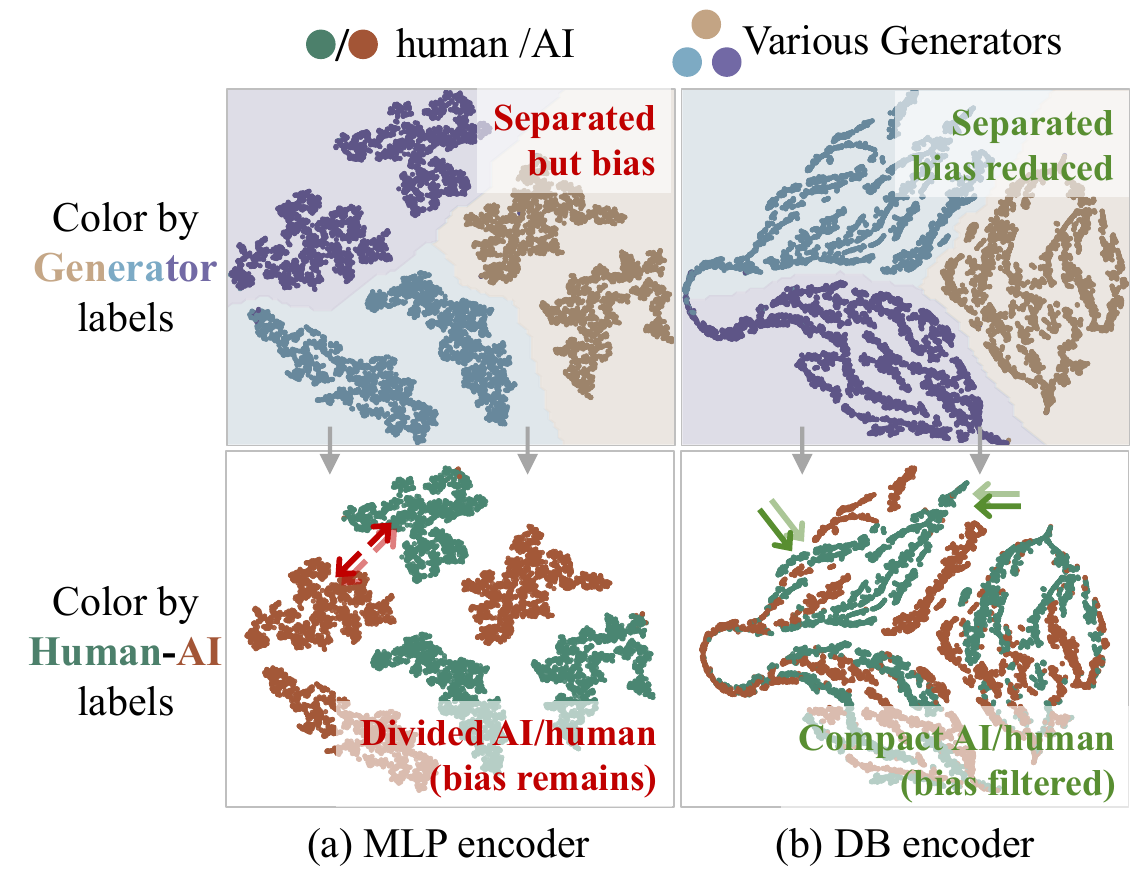} 
     \caption{MLP vs. DB encoder representations of generator-aware branch $\mathbf{g}$. With Human–AI coloring (top), both separate distinct generators, but MLP retains human-AI bias while DB filters it. 
     With generator coloring (bottom), MLP forms divided human and AI samples, whereas DB suppresses bias and yields compact, generator-aware clusters.}
    \label{fig:ib_option}
\end{figure}

Complementing the AI-detection analysis, we further examine the generator-aware branch. Figure~\ref{fig:ib_option} shows t-SNE visualizations of these representations under the same setup (GPT, BigScience, and FLAN-T5). In the MLP baseline, generator-aware features entangle with authenticity cues, implicitly encoding AI-vs-human distinctions. In contrast, our DB encoding yields clearer generator separation while suppressing unintended authenticity leakage. This confirms our disentanglement framework not only produces generator-invariant AI-detection features, but also enforces semantic purity in the generator-aware branch, strengthening overall robustness to unseen generators.

\end{document}